\documentclass[10pt,twocolumn,letterpaper]{article}

\usepackage{cvpr}              % To produce the CAMERA-READY version
\usepackage{graphicx}
\usepackage{amsmath}
\usepackage{amssymb}
\usepackage{booktabs}

\usepackage{multirow}
\usepackage{threeparttable}
\usepackage{subcaption}
\usepackage{xcolor}
\usepackage{subcaption}
\usepackage{dsfont}
\usepackage{algorithm}
\usepackage{algorithmicx}
\usepackage{algpseudocode}

\usepackage{pifont} % circled number

\newcommand{\bheading}[1]{{\noindent{\textbf{#1}}}}

\newcommand{\tabincell}[2]{\begin{tabular}{@{}#1@{}}#2\end{tabular}}

\usepackage[pagebackref,breaklinks,colorlinks]{hyperref}

\usepackage[capitalize]{cleveref}
\crefname{section}{Sec.}{Secs.}
\Crefname{section}{Section}{Sections}
\Crefname{table}{Table}{Tables}
\crefname{table}{Tab.}{Tabs.}

\def\cvprPaperID{3827} % *** Enter the CVPR Paper ID here
\def\confName{CVPR}
\def\confYear{2023}

\begin{document}

\title{Information-guided pixel augmentation for pixel-wise contrastive learning}

\author{Quan Quan\textsuperscript{1} \quad
	Qingsong Yao\textsuperscript{1}\quad
	Jun Li\textsuperscript{1}\quad
	S.Kevin Zhou$^{2,1}$ \\
	$^1$ Key Lab of Intelligent Information Processing of Chinese Academy \\ of Sciences (CAS), Institute of Computing Technology \\
	$^2$ University of Science and Technology of China \\
    % {\tt\small quanquan@miracle.ict.ac.cn, yaoqingsong19@mails.ucas.edu.cn, \\ junli@miracle.ict.ac.cn, s.kevin.zhou@gmail.com}
% 	{\tt\small \{junwei@link., qinwang1@., lizhen@\}cuhk.edu.cn, s.kevin.zhou@gmail.com}
}
\maketitle

%%%%%%%%% ABSTRACT
\begin{abstract}
   Contrastive learning (CL) is a form of self-supervised learning and has been widely used for various tasks. Different from widely studied instance-level contrastive learning, pixel-wise contrastive learning mainly helps with pixel-wise tasks such as medical landmark detection. The counterpart to an instance in instance-level CL is a pixel, along with its neighboring context, in pixel-wise CL. Aiming to build better feature representation, there is a vast literature about designing instance augmentation strategies for instance-level CL; but there is little similar work on pixel augmentation for pixel-wise CL with a pixel granularity. In this paper, we attempt to bridge this gap. We first classify a pixel into three categories, namely {low-, medium-, and high-informative}, based on the information quantity the pixel contains. Inspired by the ``InfoMin" principle, we then design separate augmentation strategies for each category in terms of augmentation intensity and sampling ratio. Extensive experiments validate that our \textbf{information-guided pixel augmentation} strategy succeeds in encoding more discriminative representations and surpassing other competitive approaches in unsupervised local feature matching. Furthermore, our pretrained model improves the performance of both one-shot and fully supervised models. To the best of our knowledge, we are the first to propose a pixel augmentation method with a pixel granularity for enhancing unsupervised pixel-wise contrastive learning.
\end{abstract}

%%%%%%%%% BODY TEXT
\section{Introduction}

Contrastive learning (CL) is a form of self-supervised learning (SSL), where data provides supervision via utilizing proxy tasks~\cite{cole2022does}. Compared with previous SSL approaches, contrastive learning focuses on learning representations of data in the embedding space by contrasting samples with the same and different distributions \cite{chen2020improved,he2020momentum,DBLP:conf/icml/ChenK0H20,chen2020big}. Currently, it has been widely applied to boost the performance of classification, a typical instance-level task. However, when applying instance-level CL for pixel-level downstream tasks~\cite{zhou2015medical,liu2010search}, {\textit e.g.}, landmark detection, segmentation, and object detection, it offers limited help when transferring instance-level contrastive learning methods to pixel-level downstream tasks directly, due to the discrepancy of supervision granularity. 
As a result, researchers plagued by pixel-level downstream tasks prefer pixel-wise CL. 

The performance of pixel-wise CL largely depends on the generation of positive and negative pairs, which motivates us to think of such a challenge: {\it how to augment the training pairs for pixel-wise CL effectively?} 
Tian {\it et al.}~\cite{tian2020makes} gives their answer for instance-level CL. They propose the so-called ``InfoMin'' principle: A good positive pair of images should share the minimal information necessary to perform well in the downstream task~\cite{tian2020makes}. Tian {\it et al.}~\cite{tian2020makes} argue that learning representations which throw out information about nuisance variables is preferable as it can improve generalization and decrease sample complexity on downstream tasks. 
Additionally, the optimal augmentation depends on the minimal shared information~\cite{tian2020makes}, which serves as inspiration and motivation for us to construct an ``optimal'' augmentation in pixel-wise CL. 
% They also demonstrate that the InfoMin principle can be practically applied by simply seeking stronger data augmentation to further reduce mutual information toward a sweet spot.   

\begin{figure}
    \centering
    \includegraphics[width=0.8\linewidth]{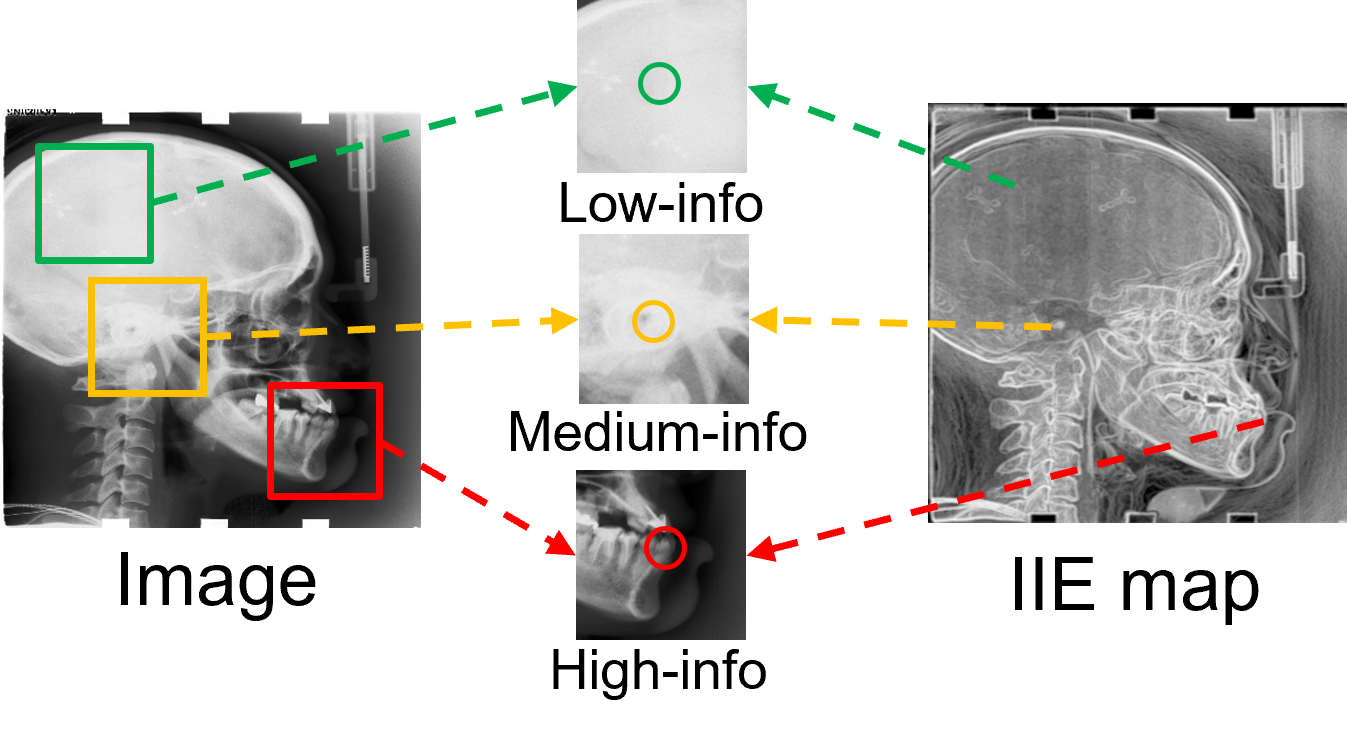}
    \caption{{\bf Main idea}: the pixels are clustered into three classes: low-, medium-, high-informative, and each class is supplemented by corresponding strategy. IIE: image information entropy.}
    \label{fig:idea}
\end{figure}

In this paper, we aim to tackle this challenge by unveiling the advantage and characteristics of pixel-wise CL, and optimizing the acquisition of positive and negative pairs \textit{with a pixel granularity} in an unsupervised style.
% To answer this question, we have to mine the reason of the advantage and characteristics of pixel-wise CL. 

First, images are signals with unbalanced information quantity at the pixel level. Each pixel with its context offers aids with varying degrees for the understanding of parts, objects and structures in an image. As a typical pixel-wise task, landmark detection usually benefits more from informative pixels.
Therefore, measuring information of pixels within an image and concentrating more on informative regions will be helpful. 
A practical solution to achieve this is to introduce the knowledge from annotations~\cite{DBLP:conf/iclr/LiuZJD22,wang2021exploring}, {\it e.g.} Liu {\it et al.}~\cite{DBLP:conf/iclr/LiuZJD22} train a semi-supervised model to generate confidence maps where the pixels with high confidence are more involved in CL training. However, few of the existing approaches address this issue in an unsupervised manner. 
To break through this wall, we introduce 
% entropy, which regards information quantity as the amount of uncertainty about a signal with a given distribution. 
image information entropy (IIE), a derived form of information entropy, to estimate the complexity of a pixel with its neighboring content. With the aid of IIE, we can divide pixels into low-, medium-, and high-info groups and compare more contrastive pixel pairs in more informative regions. 
% For an image, local entropy estimates the image from the perspective of the complexity of content, which could estimate information meanwhile. 

Second, pixels contain different amounts of semantic information related to downstream tasks, resulting in different levels of effectiveness for contrastive learning according to the ``InfoMin'' principle~\cite{tian2020makes}. Positive pairs in pixels with more semantic information will get hurt more in CL when having stronger augmentation. Stronger augmentation brings more perturbations in absolute value to positive pairs, and erodes the semantic information. 
%results in their differences in optimal augmentation parameters. 
Therefore, in this paper, we apply different strategies to different pixels: weaker augmentation for higher informative pixels and stronger augmentation for lower informative pixels. 

In sum, we propose an \textbf{information-guided pixel augmentation} strategy for pixel-wise contrastive learning as in Fig.~\ref{fig:idea}, with the following notable contributions: 
\begin{itemize}
\item To the best of our knowledge, we are \textbf{the first} to propose a pixel augmentation method \textbf{with a pixel granularity} for enhancing unsupervised pixel-wise contrastive learning. 
\item We introduce the metric of {\bf image information entropy} (IIE) to quantify the information a pixel contains. Guided by the IIE value, we then divide pixels into low-, medium-, and high-info groups and demonstrate the importance of high- and medium-info pixels in pixel-wise contrastive learning.
\item We leverage \textbf{the ``InfoMin" principle} and empirically prove its applicability to pixel-wise CL by using adaptive augmentation intensities for different groups of pixels, which leads to better experimental performances.
\item We extensively validate the effectiveness of our core strategies that include the biased selection of high-info pixels and the use of adaptive augmentation strategies, %that help conventional SSL methods~\cite{yao2021one} train a more powerful pretrained model, 
which significantly enhances the one-shot model CC2D~\cite{yao2021one} from 2.85mm to 2.31mm and 2.20mm to 1.70mm in MRE on Cephalometric and Hand X-ray datasets, respectively. Furthermore, it improves the performance of the state-of-the-art supervised landmark detection model ERE~\cite{McCouat_2022_CVPR}.
% \textcolor{red}{THESE numbers are less impressive} enhances one-shot landmark detection~\cite{yao2021one}, 
\end{itemize}

\section{Related Work}
\subsection{Contrastive learning}
% Self-supervised learning (SSL) leverages information in data itself as the supervision, providing a solution for training from unlabeled data. Among all varieties of SSL, 
Contrastive learning (CL) is one of the most powerful paradigms of self-supervised learning, leading to state-of-the-art performances in many vision tasks~\cite{chuang2020debiased,you2020graph,wang2022contrastive}. Most existing methods, like MoCo~\cite{chen2020improved, he2020momentum}, SimCLR~\cite{DBLP:conf/icml/ChenK0H20, chen2020big}, BYOL~\cite{DBLP:conf/nips/GrillSATRBDPGAP20} and BarlowTwins~\cite{DBLP:conf/icml/ZbontarJMLD21}, are designed and optimized in instance-level comparisons, benefiting the trained model with more discriminative and generalizable representations. But such instance-level modeling leads to sub-optimal representations for downstream tasks requiring pixel-level prediction, {\it e.g.}, segmentation, object detection, and landmark detection. Recently some pixel-level methods attempt to learn dense feature representations~\cite{DBLP:conf/cvpr/XieL00L021,DBLP:conf/iccv/GansbekeVGG21}. Xie {\it et al.} \cite{DBLP:conf/cvpr/XieL00L021} propose an unsupervised pixel-level contrastive learning framework and achieve better performance on dense tasks like segmentation and detection. Multi-scale pixel-wise contrastive proxy task based on InfoNCE loss~\cite{DBLP:journals/jmlr/GutmannH10, oord2018representation} is introduced in \cite{yao2021one} and achieves great performance in medical landmark detection, which is used for our feature extractor. Besides, some researchers introduce pixel-wise contrastive learning into supervised~\cite{DBLP:conf/iccv/WangZYDKG21} or semi-supervised learning~\cite{DBLP:conf/iclr/LiuZJD22,DBLP:conf/iccv/ZhongYWY0W21,DBLP:conf/ijcai/WangLL0K22,wang2021exploring,you2022bootstrapping,DBLP:conf/miccai/YouZSD22,DBLP:conf/nips/ChaitanyaEKK20,DBLP:journals/corr/abs-2112-09645,DBLP:conf/miccai/HuZXS21,DBLP:journals/corr/abs-2105-12924,zhao2021contrastive,alonso2021semi,zhou2021c3,hu2021region} and succeed to improve their models. Other tasks~\cite{you2022mine,DBLP:journals/corr/abs-2011-12640,DBLP:journals/corr/abs-2012-13089,DBLP:journals/tmi/YanCJMGHTXLL22}, such as weakly supervised methods also benefit greatly from the pixel-wise contrastive technique~\cite{du2022weakly,DBLP:conf/iclr/KeHY21}. However, to the best of our knowledge, none of them try to boost pixel-wise contrastive learning in an unsupervised manner. 

\subsection{Data augmentation}
% [1] Improving contrastive learning by visualizing feature transformation
% [2] What makes for good views for contrastive learning?
Data augmentation is the subject of extensive research today. It is also regarded as the process of generating different "views" from raw data in contrastive learning.
Nowadays, researchers strive to create diverse views on various tasks in different ways.
MixUp~\cite{zhang2018mixup,verma2019manifold,yun2019cutmix,kim2020mixco} blends images and their labels in a paired style to strengthen supervised learning methods.
Manifold MixUp~\cite{verma2019manifold} is designed for supervised learning, applying regularization on features, while Un-mix~\cite{shen2022mix} recommends MixUp in the image/pixel space for self-supervised learning; i-Mix~\cite{DBLP:conf/iclr/LeeZSLSL21} regularizes the training progress in contrastive training by mixing instances in both the input and virtual label spaces. MoChi~\cite{kalantidis2020hard} proposes mixing the negative samples in the embedding space for hard negatives augmentation to help with CL models but hurt the classification accuracy. Zhu {\it et al.}~\cite{zhu2021improving} mix both positive and negative samples to further create more views to boost CL. Besides, some other methods control the selection of views, such as Robinson {\it et al.}~\cite{robinson2020contrastive} select top-k hardest views as negative views. Some other researchers leverage the knowledge from annotations to generate views~\cite{DBLP:conf/iclr/LiuZJD22}. Liu {\it et al.}~\cite{DBLP:conf/iclr/LiuZJD22} sample positive views via confidence maps generated by semi-supervised models. Wang {\it et al.}~\cite{wang2021exploring} treat the pixels belonging to the same categories as positive pairs with the aid of segmentation masks.
In this paper, we introduce IIE to guide the generation process of positive pairs in an unsupervised style, and apply adaptive augmentation to generate more various pixels for pixel-wise CL effectively. 
% Positive features are extrapolated to increase the hardness of positives, and negative features in the memory queue are interpolated to increase the diversity. 
% MixUp for contrastive learning Mixup \cite{zhang2018mixup} and its numerous variants \cite{verma2019manifold,yun2019cutmix,kim2020mixco} provide highly effective data augmentation strategies when paired with a cross-entropy loss for supervised and semi-supervised learning.

\section{Preliminaries} \label{sec:pre}

\subsection{Contrastive learning setup} \label{sec:contrast}
Contrastive learning aims to learn representation by clustering \textit{positive pairs}, representing similar samples with the same semantic content, and discriminating \textit{negative pairs}, representing dissimilar samples with different semantic content. Specifically, the positive pairs are often obtained from views generated from the same image with different augmentations.
Similar to instance-wise CL, pixel-wise CL considers image patches at the identical location as positive pairs. 

Let $p_{data}$ be the data distribution and $p_{pos}(\cdot, \cdot)$ the distribution of positive pairs. We have \textit{contrastive loss} (InfoNCE loss), which has been shown effective by many recent contrastive learning methods~\cite{oord2018representation,yao2021one}, as follows:
\begin{equation}
\begin{aligned}
    &\mathcal{L}_\text{contrastive}(f;\tau,M) = \\
    &\underset{ \underset{\{\mathbf{x}^-_i\}^M_{i=1} \overset{\text{i.i.d}}{\sim} p_\texttt{data}}{(x,x^+) \sim p_{pos}}}{\mathbb{E}} \Big[-\log\frac{e^{f(\mathbf{x})^\top f(\mathbf{x^+}) / \tau}}{e^{f(\mathbf{x})^\top f(\mathbf{x^+}) / \tau} + \sum_{i=1}^M e^{f(\mathbf{x}_i^-)^\top f(\mathbf{x^+}) / \tau}} \Big] , \\
\end{aligned}
\label{eq:loss1}
\end{equation}
where $f$ is the feature to be learned, $\tau>0$ is a temperature hyper-parameter, and $M$ is the number of negative samples.  
% The standard form of contrastive loss in Eqn. (\ref{eq:loss1}) can be rewritten as follows:
% \begin{equation}
% \begin{aligned}
%     &\mathcal{L}_\text{contrastive}(f;\tau,M) = - \frac{1}{\tau} \underset{(x,x^+) \sim p_{pos}}{\mathbb{E}} [f(\mathbf{x})^\top f(\mathbf{x^+})] \\
%     &+\underset{ \underset{\{\mathbf{x}^-_i\}^M_{i=1} \overset{\text{i.i.d}}{\sim} p_\texttt{data}}{(x,x^+) \sim p_{pos}}}{\mathbb{E}} \Big[ \log \Big( e^{f(\mathbf{x})^\top f(\mathbf{x^+}) / \tau} + \sum_{i} e^{f(\mathbf{x}_i^-^\top )f(\mathbf{x^+}) / \tau}\Big) \Big]. 
% \end{aligned}
% \label{eq:eq2}
% \end{equation}

\subsection{Indicators} \label{sec:indicators}
%For more convincing analysis, it is rational to find some indicators to comprehend the views with statistic description. 
To better characterize pixels and their relationships, we hereby introduce three indicators: (1) Image Information Entropy (IIE); (2) {Mutual information} (MI).

{\it Image Information Entropy} (IIE). IIE is a metric introduced to quantify the information of a pixel using a small patch centered at this pixel in our method. For a specific pixel $p$, we crop a $k \times k$ patch $X_p$ whose center locates at $p$, obtain the grayscale value distribution $\mathcal{G}$, and calculate the entropy of $\mathcal{G}$ as the entropy of the pixel $p$.
\begin{equation}
% \begin{aligned}
    H(p) = H(\mathcal{G}) = - \sum_{g \in \mathcal{G}} P(g) \log P(g)
% \end{aligned}
    % H(x) = {\mathbb{E}} [\log(P(x=c_k))],
\end{equation}

{\it Mutual information} (MI). MI is a metric to assess the statistical relevance of two entities. Here we use MI to estimate the relevance of two patches of interest. In practice, similar to IIE, we first obtain the grayscale value distributions $\mathcal{G}_p$ and $\mathcal{G}_q$ of pixels $p$ and $q$ and then calculate the mutual information between $\mathcal{G}_p$ and $\mathcal{G}_q$.
\begin{equation}
%\begin{aligned}
\hat{I}[p;q] = \sum_{y \in \mathcal{G}_q} \sum_{x \in \mathcal{G}_p} P_{\mathcal{G}_p,\mathcal{G}_q} (x,y) log \Big(\frac{P_{\mathcal{G}_p,\mathcal{G}_q} (x,y)}{P_{\mathcal{G}_p} (x) P_{\mathcal{G}_q (y)}} \Big).
%\end{aligned}
\end{equation}
Compared with the contrastive loss which fluctuates along with the network training, MI is more stable and always gives a definite quantitative result for a particular pair.

\subsection{Task description} \label{sec:task}
In this study, we focus on a medical application of the local feature matching problem, used for medical landmark detection. We follow \cite{quan2021images} to build our basic encoder, which is trained by pixel-wise contrastive learning. Specifically, in the training stage, two encoders $F$ and $F'$ with the same architecture are fed with the original image $X$ and its augmented version $X'$, respectively. Next, the features from different encoders but with responses to the same corresponding pixel are matched to calculate the CL loss.

In the inference stage, we denote the set of points to match by $P=\{p_1,p_2,\dots,p_L\}$. Suppose that $p_l^T \in P^T$ is the $l^{th}$ point in the template $T$, its corresponding $p_l^X$ in the image $X$ is found by the following \textit{searching-and-maximizing} problem:
\begin{equation}
p_l^X = \arg\max_{p} ~ s[~F \circ T(p_l^T), F \circ X(p)~]; ~p_l^T \in P^T,
\label{eq:s}
\end{equation}
where $p$ is coordinates of a pixel, $s$ is a similarity function, and $F\circ X(p)$ computes the feature map for the image $X$ and then extracts the feature vector at pixel $p$. The goal of local feature matching is to accurately match points with semantic consistency. It should be noted that all analysis experiments presented below are conducted on the Cephalometric dataset~\cite{wang2016benchmark}.

\section{Method}
\begin{figure}
    \centering
    \includegraphics[width=1\linewidth]{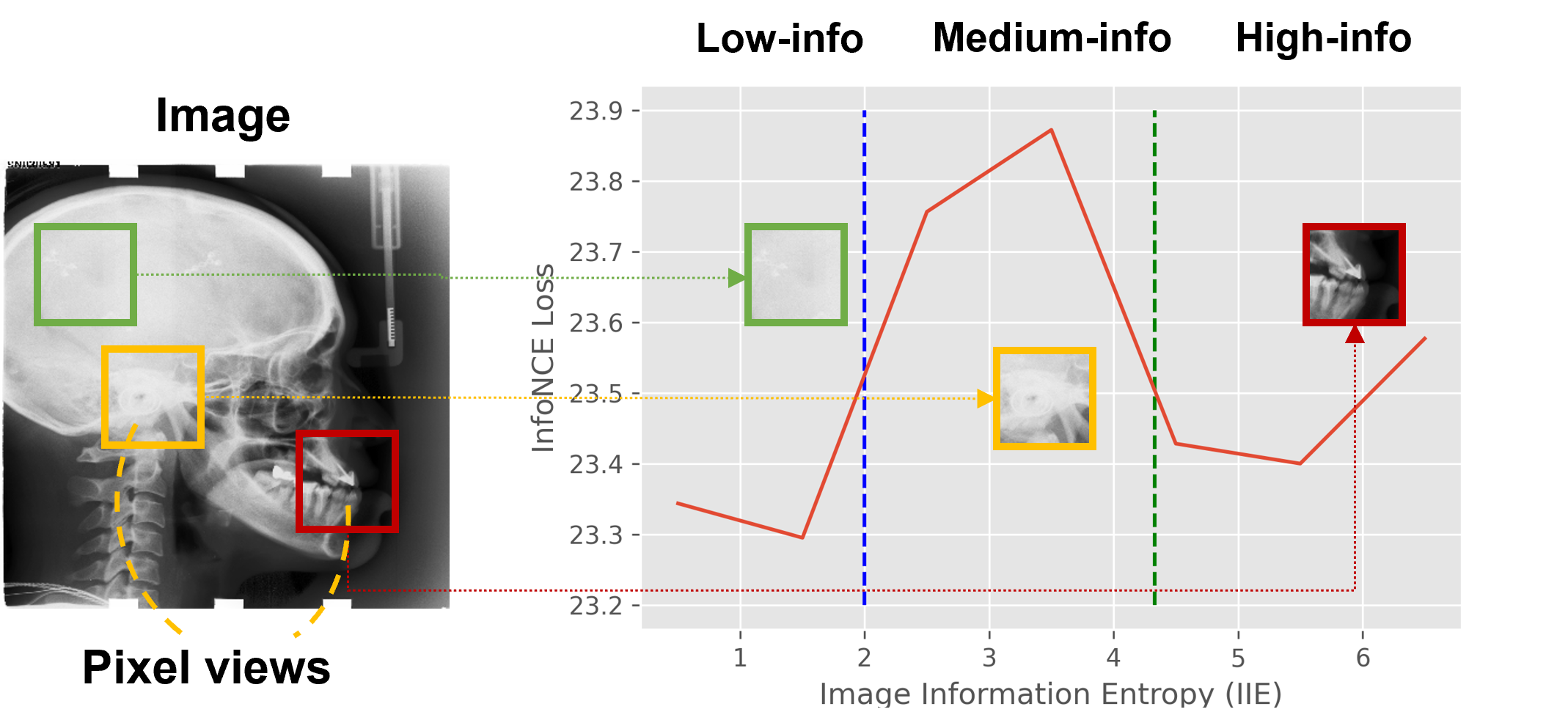}
    \caption{The curve of the contrastive loss (\textcolor{red}{red line}) against different IIE thresholds, which split pixels into three categories: low-info pixels (left of \textcolor{blue}{blue} dashed line); medium-info pixels (between \textcolor{blue}{blue} and \textcolor{green}{green} dashed lines); high-info pixels (right of \textcolor{green}{green} dashed line).}
    \label{fig:loss_entr}
\end{figure}

Pixels behave differently as shown in Figure~\ref{fig:loss_entr}, inspiring us to treat them with different strategies for each pixel. We naturally reform the standard contrastive loss into a pixel-wise adaptive loss:
\begin{equation}
    \mathcal{L} = \sum_i \mathcal{L}_i(\rho_i, A_i) = \sum_i \rho_i \mathcal{L}(\tau(x_i, A_i), x_i),
\end{equation}
where $x_i$ refers to a pixel (along with its content in image patch), $\tau(x_i, A_i)$ refers to transformation on $x_i$ with augmentation parameters $A_i$ and $\rho_i$ refers to the weights of $x_i$ during training. However, the above ideal case of customizing strategies $\{\rho_i, A_i\}$ for each pixel $x_i$ is hard to achieve due to its time-consuming nature, thus a practical way is to cluster $x_i$ into several classes and adjust $\{\rho, A \}$ for each class. 

\subsection{Information-guided pixel categorization} \label{sec:lowinfo}
%In this section, we discuss the details of low informative pixels and propose a strategy: reduce the proportion of low informative entropy pixels and increase others. Following the above discussion, we know different pixels have different quantity of information.
% To research the traits of pixels in specific information ranges, two indicators are introduced to quantize their characteristics: 
To cluster $x_i$, we use {\it Image Information Entropy} (IIE) and {\it contrastive loss} (Loss), which are already detailed in section~\ref{sec:pre}. Figure~\ref{fig:loss_entr} shows a U-shaped curve of the loss against IIE, nicely splitting pixels into three categories: low-, medium- and high-info pixels. We list their details as follows.
\begin{enumerate}
    % \item (0) Original views: original views are those images collected from real world or specific environments without any additional operations in texture or semantic level. 
    \item Low-info pixels featuring low IIE and low loss: most of them contain less semantic information and much noise (low IIE) but are easy to discriminate (low loss).
    \item Medium-info pixels featuring medium IIE and high loss: they have medium texture information (medium IIE) but are hard to discriminate for networks (high loss).
    \item High-info pixels featuring high IIE and low loss: most of them are edges and corners with much deterministic information (high IIE) and they are easy to discriminate for networks (low loss).
\end{enumerate}

With the division of all pixels into three categories, our pixel augmentation goal can be formulated as follows:
\begin{equation}
\begin{aligned}
    \mathcal{L}(\rho_l,\rho_m,&\rho_h, A_l, A_m, A_h) = \\
    &\sum_{\alpha \in \{l,m,h\}} \rho_\alpha \mathcal{L}_\alpha(f(\tau (x_\alpha, A_\alpha), x_\alpha)),
\end{aligned} 
\label{eq:L}
\end{equation}
where the total loss $\mathcal{L}$ is split into 3 losses $\mathcal{L}_l$, $\mathcal{L}_m$ and $\mathcal{L}_h$ with their sampling ratios $\rho_l$, $\rho_m$, and $\rho_h$ and augmentation intensities $A_l$, $A_m$, and $A_h$ for the low-, medium- and high-info pixels $x_l$, $x_m$, and $x_h$, respectively. The proposed strategy of augmenting pixels is to find the optimal {\bf sampling ratio} and {\bf augmentation intensity} for each category such that the overall loss is minimized. Next, we will elaborate how to do so one by one.

\begin{figure}
    \centering
    \includegraphics[width=0.8\linewidth]{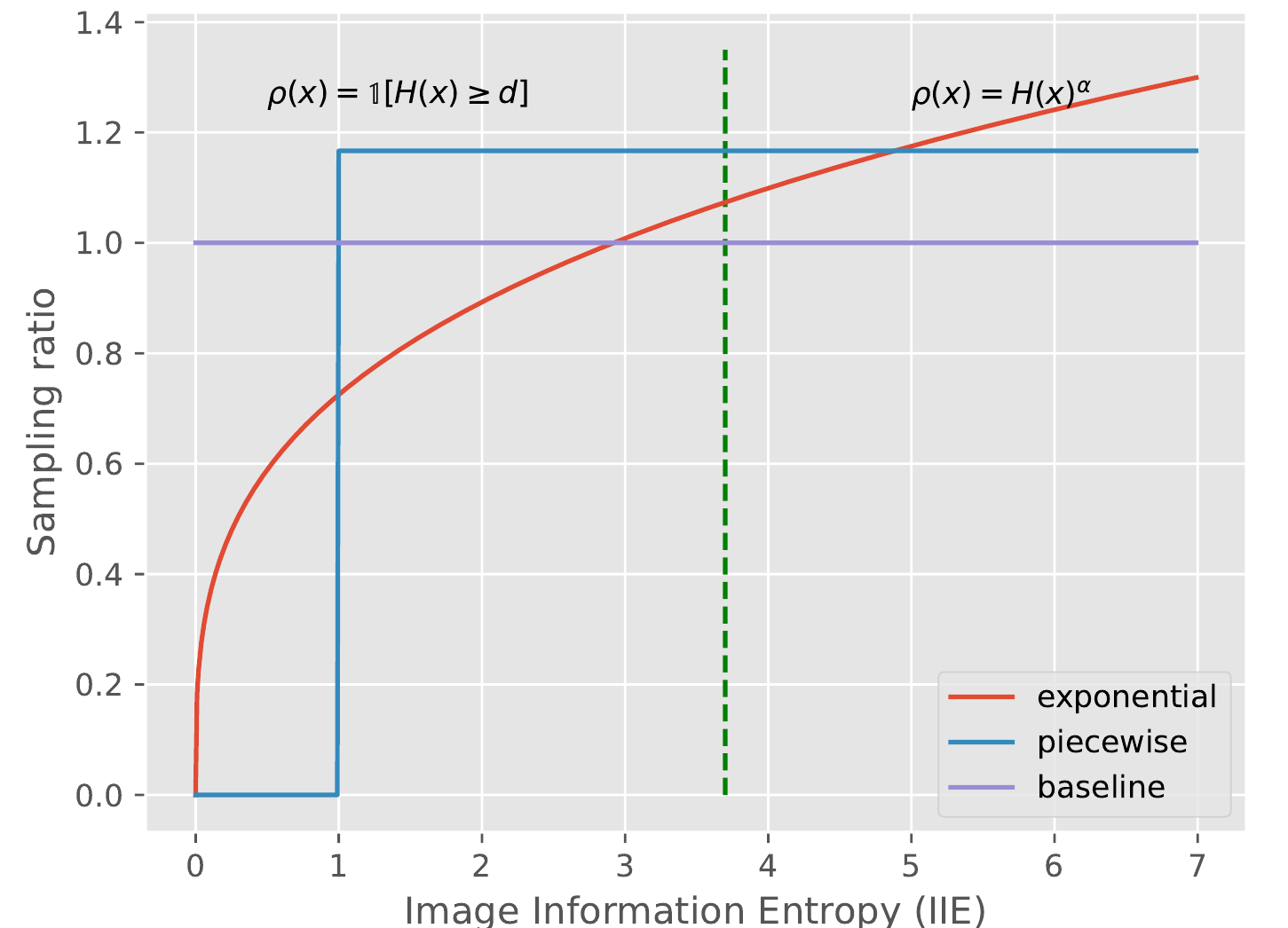}
    \caption{The \textcolor{green}{green} dashed line: entropy of all annotated landmarks are above this line. The \textcolor{violet}{violet} line (baseline): sampling probability is same for all pixels. The \textcolor{red}{red} line: sampling probability of an entropy-based sampling weight map. \textcolor{blue}{Blue} line: sampling probability of a piecewise weight map.}
    \label{fig:entr1}
\end{figure}

\begin{table}[]
    \centering
    \begin{tabular}{c|l|l|l|l|l|l}
    \multicolumn{7}{c}{Exponential weight map} \\
    \hline
        $\gamma$   &  0   & 0.01 & 0.1 & 0.3   & 0.5 & 1.0 \\
        \hline
        MRE & 2.91 & 2.76 & 2.48 & 2.46 & 2.50 & 2.51 \\
        \hline
        \hline
        \multicolumn{7}{c}{Piecewise weight map} \\
        \hline
        $d$ & 0  & 1    &  2  & 3   &  3.7 & 4.5 \\
        \hline
        MRE & 2.91 & 2.46 & 2.46 & 2.50 & 4.25 & 4.23 \\
    \hline
    \end{tabular}
    \caption{Accuracy under different weight maps.}
    \label{table:reduce_noise}
\end{table}

\subsection{Sampling ratio}  \label{sec:sratio}
Per prior knowledge, medical landmarks, usually are localized at those meaningful points to guide clinical analysis; thus, it is natural to focus on high-info regions. In addition to this intuitive motivation, we also find that the loss of low-info pixels quickly reaches a low value and remains low even when the sampling rate of low-info pixels is reduced. 
Based on both intuitive and experimental findings, we propose to reduce $\rho_l$, thereby forcing the network to focus on medium- and high-info pixels.

When reducing $\rho_l$, there comes a natural question: \textit{whether low-info areas are at all needed?}
To explore it, we design two experimental schemes to analyze the pixels: (1) exponential weight map, $\rho(x) = H(x)^\gamma$; (2) piecewise weight map, $\rho(x) = \mathds{1}[H(x) \geq d]$. For (1), we gradually increase the exponent $\gamma$ of entropy weight map; For (2), we discard low-info pixel views $x_l$ and gradually increase the threshold $d$ between $x_l$ and $x_m$ (Figure~\ref{fig:entr1}). We empirically find that discarding a large amount of low-info pixel views is not advisable. As shown in Table~\ref{table:reduce_noise}, the pixels whose entropy is below 1 contain much less information and are actually useless for our model; the pixels whose entropy is between 1 and 3 contain both useful and useless information; and the pixels whose entropy is greater than 3 are mostly useful. 
Considering the prior knowledge that the IIE values of ground truth landmarks in Cephalometric are almost greater than 3.7 (\textcolor{green}{Green} dashed line in Figure~\ref{fig:entr1}), some low-info pixels are kept to utilize their contributions to the variety of contrastive learning as they may contain the knowledge we need.  

\begin{figure}[t]
    \centering
    \includegraphics[width=\linewidth]{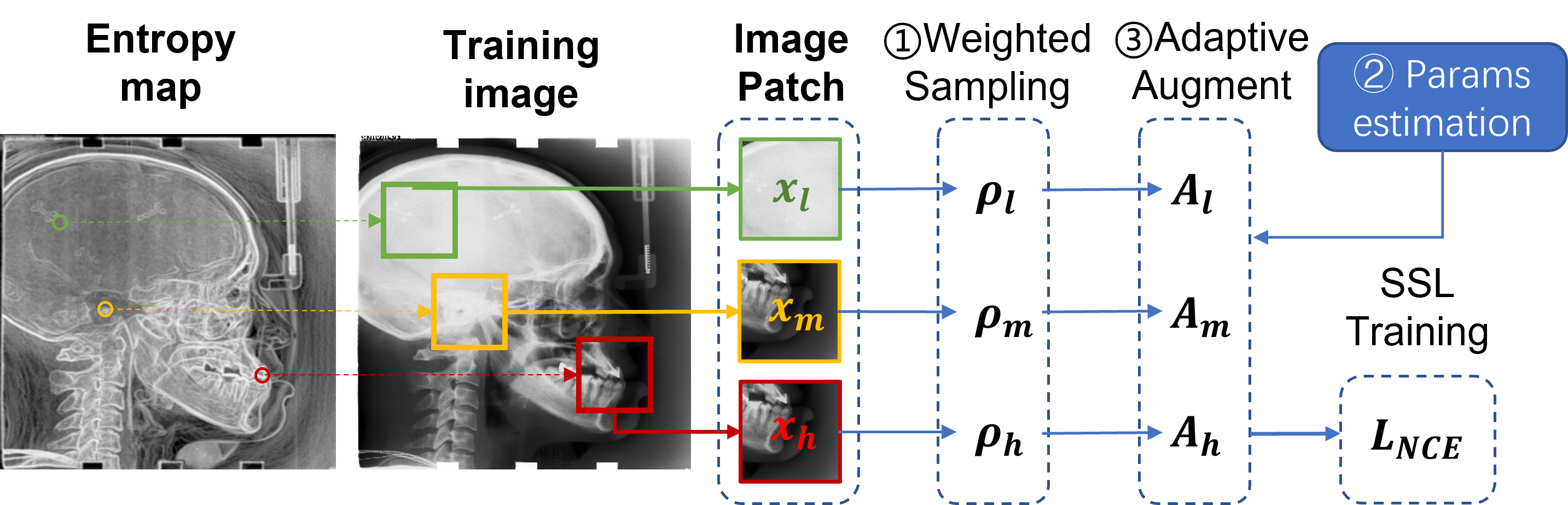}
    \caption{Pipeline of our method, consisting of 3 steps. }
    \label{fig:pipeline}
\end{figure}

\subsection{Augmentation intensity} \label{sec:info_view}

After solving $\rho$ in Eqn.~(\ref{eq:L}), we then optimize $A$ for each group of pixels by leveraging the ``InfoMin" principle.
% We aim at augmenting pixels so that their feature representation are best discovered through pixel-wise contrastive learning.

\subsubsection{Minimal necessary shared information} \label{sec:aug_int}
To investigate optimal views for contrastive learning, Tian {\it et al.} \cite{tian2020makes} propose the ``InfoMin" principle --- {\it A good set of views are those that share the minimal information necessary to perform well at the downstream task} and give two definitions and a proposition as follows:

Given two random variables $\mathbf{v_1}$ and $\mathbf{v_2}$, which are two views of the data $x$ in practice, two encoders ($f_1$ for $\mathbf{v_1}$ and $f_2$ for $\mathbf{v_2}$) resulting representations $\mathbf{z_1} = f_1(\mathbf{v_1})$ and $\mathbf{z_1} = f_2(\mathbf{v_2})$, we have\\ 
{\bf Definition 1}. {\it (Sufficient Encoder) The encoder $f_1$ of $\mathbf{v_1}$ is sufficient in the contrastive learning framework if and only if $I[\mathbf{v_1};\mathbf{v_2}]=I[f_1(\mathbf{v_1}); \mathbf{v_2}]$.} $I[\mathbf{v_1};\mathbf{v_2}]$ denotes the mutual information between $\mathbf{v_1}$ and $\mathbf{v_2}$. Generally speaking, a sufficient encoder refers to a well-trained encoder that can encode all necessary information relevant to tasks. In other words, the encoder $f_1$ is sufficient if $\mathbf{z_1}$ has kept all necessary information that the contrastive learning objective requires. Symmetrically, $f_2$ is sufficient if $I[\mathbf{v_1}; \mathbf{v_2}] = I[\mathbf{v_1}; f_2(\mathbf{v_2})]$.

\noindent {\bf Definition 2}. {(\it Minimal Sufficient Encoder) A sufficient encoder $f_1$ of $\mathbf{v_1}$ is minimal if and only if $I[f_1(\mathbf{v_1}); \mathbf{v_1}] \leq I[f(\mathbf{v_1}); \mathbf{v_1}]$, $\forall f$ that is sufficient.} 
Among those encoders that are sufficient, the minimal ones, that only extract relevant information of the contrastive task and throw away other irrelevant information, are proved to be the most robust encoders, and also what we mainly care about.

\noindent\textbf{Proposition 1.} \textit{Suppose $f_1$ and $f_2$ are minimal sufficient encoders. Given a downstream task $\mathcal{T}$ with label $\mathbf{y}$, the optimal views created from the data $\mathbf{x}$ are $(\mathbf{v_1}^*, \mathbf{v_2}^*) = \arg\min_{\mathbf{v_1}, \mathbf{v_2}} I[\mathbf{v_1};\mathbf{v_2}]$, subject to $I[\mathbf{v_1}; \mathbf{y}]=I[\mathbf{v_2};\mathbf{y}]=I[\mathbf{x};\mathbf{y}]$. Given $\mathbf{v_1}^*$, $\mathbf{v_2}^*$, the representation $\mathbf{z_1}^*$ (or $\mathbf{z_2}^*$) learned by contrastive learning is optimal for $\mathcal{T}$, thanks to the minimality and sufficiency of $f_1$ and $f_2$.}
The above proposition states the importance of the pairs with minimal necessary shared information, which motivates us to find them in pixel-wise CL.

\begin{figure}
    \centering
    \includegraphics[width=0.9\linewidth]{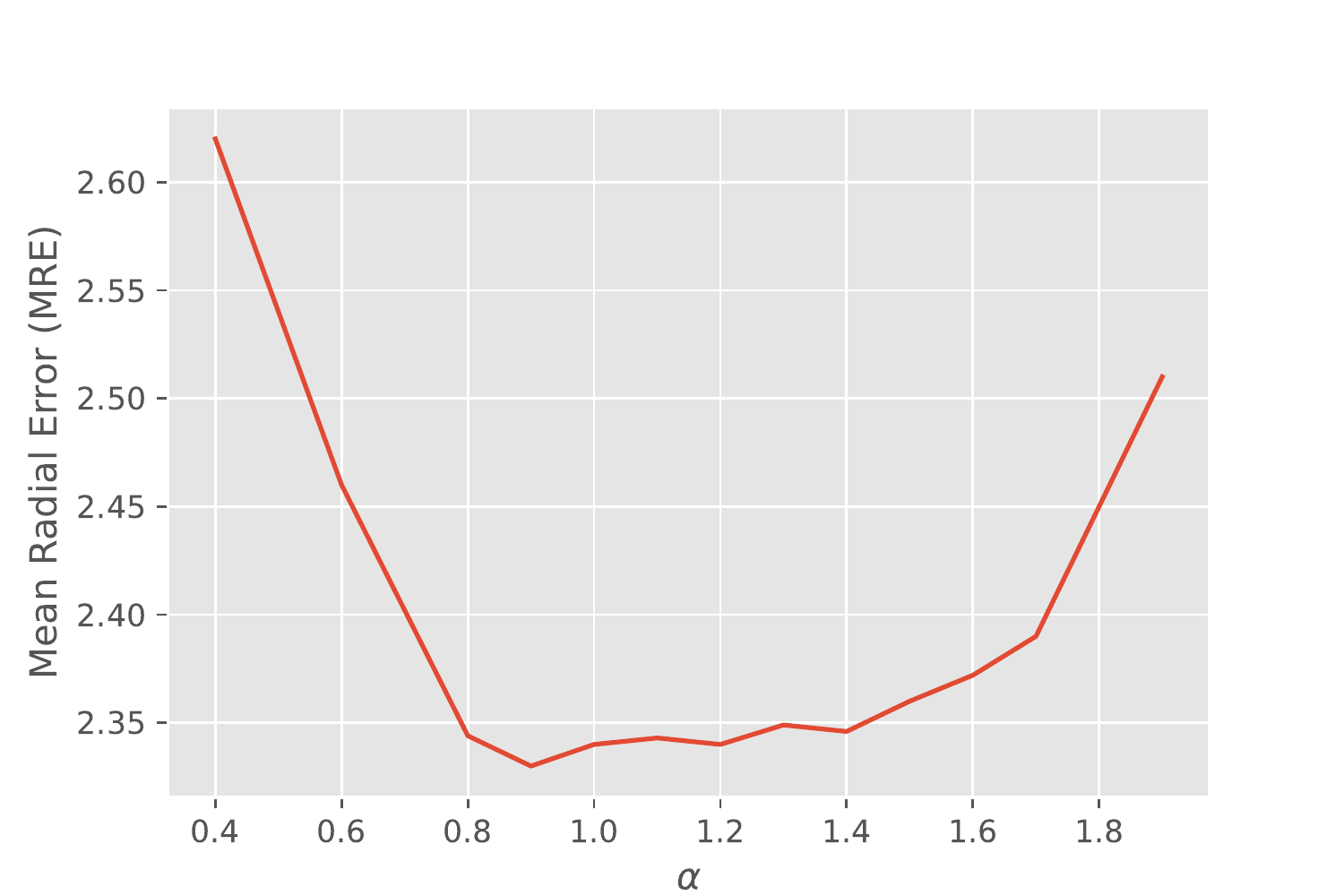}
    \caption{Model performance with different $\hat{\alpha}$.}
    \label{fig:alpha}
\end{figure}

\subsubsection{Augmentation intensity estimation}
Inspired by the ``InfoMin" principle, we can estimate our optimal augmentation parameters by minimizing the gap between the MI among augmented positive pairs and the minimal necessary MI among samples of the same class. 
% Inspired by the ``InfoMin" principle, we can estimate our optimal augmentation parameters by maximizing augmentation intensity while keeping necessary minimal mutual information among corresponding pixels of the same semantic information.  
% First, our goal is to find the minimal necessary MI which can be derived from \textbf{Proposition 1}, 
\begin{equation}
    A = \arg \min_{A \in \mathbf{A} } ||~ \underbrace{\min \{I[\mathbf{v}; \mathbf{v'}] \}}_{\textbf{(a)}} - \underbrace{\mathbb{E}\{\hat{I}[\mathbf{v};\tau(\mathbf{v},A)] \}}_{\textbf{(b)}}||_1 
    \label{eq:A}
\end{equation}
Two components are involved in Eqn.~(\ref{eq:A}): (a) estimating minimal necessary MI; (b) estimating MI among augmented pairs.

\bheading{Approximating minimal necessary MI:}
Due to the difficulty of directly obtaining the minimal necessary MI, an upper bound is proposed to approximate it as follows:
\begin{equation}
\begin{aligned}
    \min I[\mathbf{v}; \mathbf{v'}] &= \min \alpha I[\mathcal{G}(\mathbf{v}); \mathcal{G}(\mathbf{v'})] \\
    &\leq \mathbb{E} \{\alpha I[\mathcal{G}(\mathbf{v}); \mathcal{G}(\mathbf{v'})]\} \\
    &\leq \hat{\alpha} \mathbb{E} \{\hat{I}[\mathbf{v}; \mathbf{v'}]\};~ \hat{\alpha}=\max(\alpha),
\end{aligned}
\end{equation}
where $\mathbf{v}$ and $\mathbf{v'}$ are views corresponding to an identical key point), $\mathcal{G}(\mathbf{v})$ is the grayscale value distribution of $\mathbf{v}$, and $\alpha$ denotes a scale factor. As it is hard to estimate the mutual information between images, the common alternative is to estimate the MI between the grayscale value distributions of images. However, the progression from $\mathbf{v}$ to $\mathcal{G}(\mathbf{v})$ loses information due to dimensional reduction; we regard $\alpha$ as the inverse ratio of information loss.  

With this approximation, Eqn~.(\ref{eq:A}) can be rewritten into 
\begin{equation}
    A = \arg \min_{A \in \mathbf{A} } \mathbb{E} \Big\{  || \hat{\alpha} \hat{I}[\mathbf{v};\mathbf{v'}] - \hat{I}[\mathbf{v};\tau(\mathbf{v}, A)] ||_1 \Big\}    
\end{equation}
% where $\hat{\alpha}$ are considered as a hyper-parameter. 

\bheading{Details:} Specifically, as demonstrated in Eqn.~(\ref{eq:estimate1}), we select $k$ pixels as key points $l$ from a randomly selected image $X_0$ in a particular group (low-, medium- or high-info). We calculate the necessary mutual information of key points of interest $l_j$ by predicting the corresponding pixels $l^i_j$ in each image $X_i$ via a pre-trained encoder and then calculating mutual information between the corresponding patches located at key points. In addition, $\hat{\alpha}$ is considered as a hyper-parameter which will be analysed in ablation study.
Finally, we can adjust the augmentation parameters $A$ by increasing $A$ iteratively to achieve our goal.

\begin{equation}
    A = \arg\min_{A \in \mathbf{A}} \frac{1}{k} \Big\{ \Big|\Big| \frac{\hat{\alpha}}{n} \sum_{j=1}^k \sum_{i=1}^n I[l_j; l^i_j] - \sum_j^k I[l_j;\tau(l_j, A)] \Big|\Big|_1  \Big\}. 
    \label{eq:estimate1}
\end{equation}
% \begin{equation}
%     \mu = \mathbb{E} [I(\mathcal{G}(\mathbf{v}), \mathcal{G}(\mathbf{v'}))] = \frac{1}{k} \frac{1}{n} \sum_{j=1}^k \sum_{i=1}^n I[l_j; l^i_j].
%     \label{eq:estimate1}
% \end{equation}
% Considering the noise involved and the desire to obtain robust representations, we assume $\alpha\mu$ is the minimal necessary mutual information and set $\alpha=1$.
% Thus, we adjust our augmentation policy and parameters $A=\{a_0, a_1, ..., a_i\}$ to augment the key points $l$ and control the mutual information between augmented landmarks and the original in the range of $\mu$. By doing this, we can increase the variance of pixel views while keeping the necessary semantic information.

% \begin{equation}
%     A = \arg\min_{A \in \mathbf{A}} || \frac{1}{k} \sum_j^k I[\tau(l_j)] - \mu  ||
% \end{equation}

\begin{figure}
    \centering
      \includegraphics[width=0.9\linewidth]{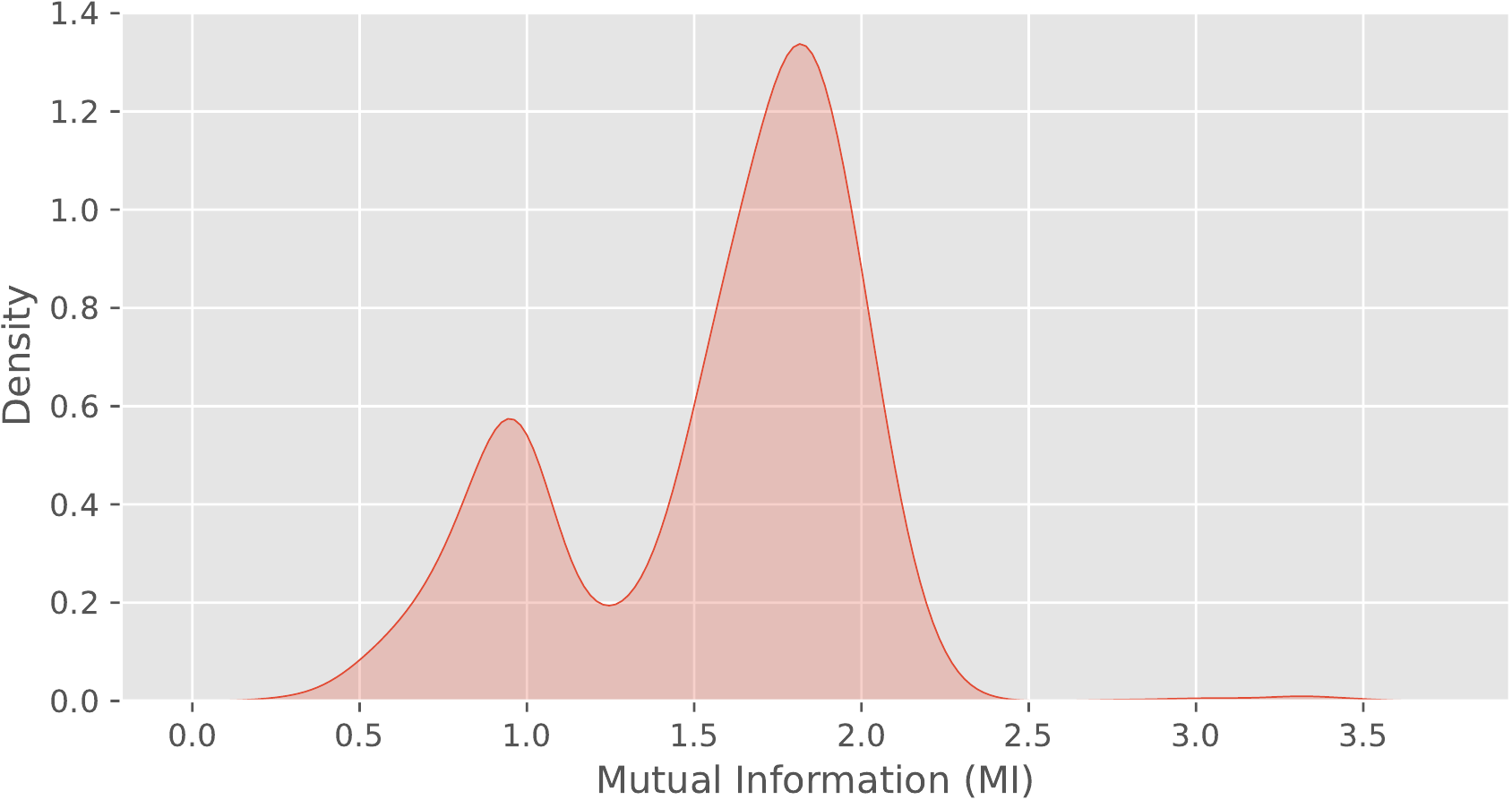}  
    \caption{Probability density of mutual information among high-info pixels.}
    \label{fig:mi_aug_bar}
\end{figure}

% \begin{equation}
%     A = \frac{1}{k} \sum_{j}^k E_s(l_j; \mu),
%     \label{eq:estimate2}
% \end{equation}
% where $E_s$ is an iterative function detailed in Algorithm~\ref{alg:a3}.
The distribution of $\hat{I}[l_j; l^i_j]$ is visualized in Figure~\ref{fig:mi_aug_bar}. 
In practice, we estimate parameters for two augmentation operations that adjust image brightness and contrast, that is, $A = \{a_{br}, a_{ct}\}$. 
Based on the above, we can obtain augmentation parameters for all three groups, $A_l$, $A_m$, and $A_h$.

\bheading{Adaptive augmentation:}
After obtaining $A_l$, $A_m$, $A_h$, we can easily apply adaptive augmentation via Eqn.~(\ref{eq:L}).

\begin{table*}
\begin{minipage}[t]{0.49\textwidth}
% \begin{table}[t]
\centering
\footnotesize    
\caption{Comparison of SSL for \textit{local feature matching} with other approaches on Cephalometric~\cite{wang2016benchmark} testset. }
\begin{threeparttable}
\begin{tabular}{l|lcccc}
\hline
 \multirow{2}{*}{Method} & \multirow{2}{*}{\tabincell{c}{MRE ($\downarrow$) \\ (mm)}} &  \multicolumn{4}{c}{SDR ($\uparrow$) (\%)} \\ \cline{3-6}
  &   & 2mm & 2.5mm & 3mm & 4mm \\ \hline
% \cline{2-7}
baseline   & 2.91 & 39.85 & 48.94 & 58.87 & 72.31 \\
% \hline
Zhu.~\cite{zhu2021improving}*   & 4.19 \textcolor{red}{(+1.28)} & 24.35 & 32.38 & 42.66 & 60.10\\
MOCHI\cite{kalantidis2020hard}*  & 2.56 \textcolor{blue}{(-0.35)} & 50.35 & 59.19 & 68.70 & 82.10 \\
Un-mix\cite{shen2022mix}*  & 2.54 \textcolor{blue}{(-0.37)} & 44.82 & 56.48 & 67.57 & 83.87 \\
Ours   & \textbf{2.34 \textcolor{blue}{(-0.57)}} & \textbf{51.28} & \textbf{62.88} & \textbf{73.70} & \textbf{86.50} \\
\hline
\end{tabular}
\begin{tablenotes}
    \footnotesize
    \item[] * using augmentation parameters estimated for high-info pixels by our method.
\end{tablenotes}
\end{threeparttable}
\label{table:main2}
% \end{table}
\end{minipage}
\begin{minipage}[t]{0.47\textwidth}
% \begin{table}[t]
\centering
\footnotesize    
\caption{Ablation study for components in our method on Cephalometric~\cite{wang2016benchmark} testset.}
\begin{threeparttable}
\begin{tabular}{ll|lllll}
\hline
\multirow{2}{*}{Entr}  & \multirow{2}{*}{Aug} & \multirow{2}{*}{\tabincell{c}{MRE ($\downarrow$) \\ (mm)}} & \multicolumn{4}{c}{SDR ($\uparrow$) (\%)} \\ 
\cline{4-7}
&           & &     2mm & 2.5mm & 3mm & 4mm \\
\hline
            &              &   2.91 & 39.85 & 48.94 & 58.87 & 72.31 \\
\checkmark &               &   2.46 \textcolor{blue}{(-0.45)}  & 43.72 & 54.65 & 66.48 & 82.65  \\
            &\checkmark & 2.41 \textcolor{blue}{(-0.50)} & 50.08& 60.96 & 71.81 & 84.58 \\
% &\checkmark & & 2.55 \textcolor{blue}{(-0.36)} & 47.47& 58.23 & 68.84 & 82.63 \\
% \checkmark & \checkmark   &               &   2.40 \textcolor{blue}{(-0.51)}  & 48.65 & 59.91 & 71.64 & 85.85 \\ %v2_br_ct
\checkmark &  \checkmark   & \textbf{2.34 \textcolor{blue}{(-0.57)}} & \textbf{51.28} & \textbf{62.88} & \textbf{73.70} & \textbf{86.50} \\
\hline
\end{tabular}
\begin{tablenotes}
    \footnotesize
    \item[] \textbf{Entr}: entropy-based weight map; \textbf{Aug}: augmentation parameters estimation and adaptive augmentation.
\end{tablenotes}
\end{threeparttable}

\label{table:abla}
% \end{table}
\end{minipage}
\end{table*}

% \begin{algorithm}[t]
%     \caption{Iterative parameter estimation}
%     \textbf{Input:} Key point $l$, minimal necessary MI $\mu$
    
%     \textbf{Output:} Augmentation parameters $A$.
    
%     \begin{algorithmic}[1] 
%         \Function {$E_s$}{$l, \mu$}
%             \State $A_0 = \{a_{0}, a_{1}, ..., a_i\}$
%             \State $\beta = \{\beta_0, \beta_1, ..., \beta_i\}$
%             %for
%             \For{$i$ in $\{1, 2, ..., n\}$}   
%             \State $A_i = A_{i-1} + \beta A_0$
%             \State $mi = I(\tau(l, A_i), l)$
%             \If {$mi < \mu$} \State\Return $A_{i-1}$ \EndIf
%             \EndFor
%         \EndFunction
%     \end{algorithmic}
%     \label{alg:a3}
% \end{algorithm}

\begin{table*}[]
\centering
\small
\caption{Comparison of the \textit{supervised} and \textit{one-shot} approaches with different pre-trained model on Cephalometric~\cite{wang2016benchmark}, Hand X-ray~\cite{ref_scn} and H\&N 3D test sets. The {\bf best} and the \underline{second-best} performances are highlighted.}
\begin{threeparttable}
\begin{tabular}{l|l|l|lllll}
\multicolumn{8}{c}{Cephalometric Test1+2} \\ 
\hline
\multirow{2}{*}{Type} & \multirow{2}{*}{Model} & \multirow{2}{*}{Pretrain} & \multirow{2}{*}{\tabincell{c}{MRE ($\downarrow$) \\ (mm)}} &  \multicolumn{4}{c}{SDR ($\uparrow$) (\%)} \\ \cline{5-8}
 &  & & & 2 mm & 2.5 mm & 3 mm & 4 mm \\ \hline
% & Ibragimov \textit{et al.}~\cite{ibragimov2015computerized}* & 150 & - & 68.13 & 74.63 & 79.77 & 86.87\\
%  & Lindner \textit{et al.}~\cite{lindner2015fully}* & 150 & 1.77 & 70.65 & 76.93 & 82.17 & 89.85\\
%  & Urschler \textit{et al.}~\cite{ref_urschler}* & 150 & - & 70.21 & 76.95 & 82.08 & 89.01\\
\multirow{15}{*}{Supervised}  & \multirow{5}{*}{U-Net~\cite{unet}} & Imagenet~\cite{DBLP:conf/cvpr/DengDSLL009} & 2.07 & 56.54 & 67.83 & 79.43 & 91.76 \\
 &  & Zhu~\cite{zhu2021improving}  &  1.96  \textcolor{blue}{(-0.11)} & 61.38  & 72.06  & 82.02  & 93.05 \\
 &  & MOCHI~\cite{kalantidis2020hard} & \underline{1.94} \textcolor{blue}{(-0.13)} & \textbf{62.03}  & \textbf{73.51}  & \textbf{84.02}  & \underline{93.18} \\
 & & Un-mix~\cite{shen2022mix} & 1.94 \textcolor{blue}{(-0.13)} & \underline{61.95}  & \underline{73.20}  & 82.96  & 93.17  \\
 & & ours  & \textbf{1.93 \textcolor{blue}{(-0.14)}} & 61.51  & 72.84  & \underline{83.05}  & \textbf{93.38} \\
\cline{2-8}
 & \multirow{5}{*}{ERE~\cite{McCouat_2022_CVPR}$^*$} & Imagenet~\cite{DBLP:conf/cvpr/DengDSLL009} & 1.42 & \underline{80.10} & \textbf{87.34} & \textbf{91.81} & \textbf{96.19} \\
 & & Zhu~\cite{zhu2021improving} & 1.41 \textcolor{blue}{(-0.01)} & 79.24  & 86.33  & 90.92  & 95.38  \\
 & & MOCHI~\cite{kalantidis2020hard} & \underline{1.40} \textcolor{blue}{(-0.02)} & 79.98  & 86.95  & 91.14  & \underline{95.90} \\
 & & Un-mix~\cite{shen2022mix}  & 1.41 \textcolor{blue}{(-0.01)} & 79.70  & 86.68  & 91.06  & 95.48  \\
 & & ours  & \textbf{1.39 \textcolor{blue}{(-0.03)}} & \textbf{80.15}  & \underline{87.17}  & \underline{91.32}  & 95.89 \\
\cline{2-8}
%  & GU$^2$-Net~\cite{zhu2021you} & Imagenet~\cite{DBLP:conf/cvpr/DengDSLL009} & 1.69 & 76.95 & 83.98 & 88.82 & 94.21\\
 & \multirow{5}{*}{GU$^2$-Net~\cite{zhu2021you}$^*$} & None & 1.34 & 83.01 & 88.82 & 93.09 & 97.34\\
 & & Zhu~\cite{zhu2021improving} & 1.32 \textcolor{blue}{(-0.02)} & 83.26  & 89.01  & 93.04  & \underline{97.42} \\
 & & MOCHI~\cite{kalantidis2020hard} & 1.31 \textcolor{blue}{(-0.03)} & \underline{83.47}  & 89.13  & 93.07  & \textbf{97.51} \\
 & & Un-mix~\cite{shen2022mix}  & 1.31 \textcolor{blue}{(-0.03)} & 83.45  & \underline{89.47}  & \underline{93.30}  & 97.13 \\
 & & ours  & \textbf{1.28 \textcolor{blue}{(-0.06)}} & \textbf{84.50}  & \textbf{89.78}  & \textbf{93.38}  & 97.34  \\
\cline{1-8}

\multirow{5}{*}{One-shot} & \multirow{5}{*}{CC2D-S2~\cite{yao2021one}$^\#$} & Imagenet~\cite{DBLP:conf/cvpr/DengDSLL009} & 2.85 & 39.30 & 50.12 & 61.30 & 77.93 \\
 & & Zhu~\cite{zhu2021improving}  & 2.87 \textcolor{red}{(+0.02)} & 38.73  & 49.57  & 61.15  & 78.10 \\
 & & MOCHI~\cite{kalantidis2020hard}   & 2.47 \textcolor{blue}{(-0.38)} & 46.06  & 57.74  & 70.14  & 86.33  \\
  & & Un-mix~\cite{shen2022mix} & \underline{2.44} \textcolor{blue}{(-0.41)}  &  \underline{49.81}  &  \underline{60.92}  &  \underline{71.70}  &  \underline{86.35}  \\
& & ours  & \textbf{2.31 \textcolor{blue}{(-0.54)}} & \textbf{52.10}  & \textbf{63.20}  & \textbf{73.45}  & \textbf{86.73}  \\
\hline
\hline
% 2.4404525045402976,49.810526315789474,60.92631578947368,71.70526315789473,86.3578947368421

\multicolumn{8}{c}{Hand X-ray Testset} \\ 
\hline
\multirow{15}{*}{Supervised} & \multirow{5}{*}{U-Net~\cite{unet}} & Imagenet~\cite{DBLP:conf/cvpr/DengDSLL009} & 1.32 & 82.78 & 90.52 & 94.49 & 97.82 \\
 & & Zhu~\cite{zhu2021improving} & 1.30 \textcolor{blue}{(-0.02)} & 81.94  & 89.52  & 93.96  & 97.90  \\
 &  & MOCHI~\cite{kalantidis2020hard} & 1.27 \textcolor{blue}{(-0.05)} & \textbf{85.29}  & \textbf{91.63}  & \textbf{95.11}  & 98.08 \\
 &  & Un-mix~\cite{shen2022mix}  &  \underline{1.29} \textcolor{blue}{(-0.03)} & 83.88  & 91.10  & 94.97  &  \underline{98.15} \\
 & & ours  & \textbf{1.24 \textcolor{blue}{(-0.08)}} &  \underline{84.48}  &  \underline{91.33}  &  \underline{95.09}  & \textbf{98.36} \\
\cline{2-8}
 & \multirow{5}{*}{ERE~\cite{McCouat_2022_CVPR}$^*$} & Imagenet~\cite{DBLP:conf/cvpr/DengDSLL009} & 0.396 & 99.01 & \textbf{99.44} & \textbf{99.64} & \textbf{99.74} \\
 & & Zhu~\cite{zhu2021improving} & 0.381 \textcolor{blue}{(-0.015)} & 99.01  & 99.38  & 99.58  & 99.70  \\
 & & MOCHI~\cite{kalantidis2020hard} & 0.380 \textcolor{blue}{(-0.016)} & 99.02  & 99.41  & 99.56  & 95.71 \\
 & & Un-mix~\cite{shen2022mix}  &  \underline{0.380} \textcolor{blue}{(-0.016)} & 99.02  & 99.42  & 99.55  & 95.70  \\
 & & ours  & \textbf{0.377 \textcolor{blue}{(-0.019)}} & \textbf{99.04}  &  \underline{99.43}  &  \underline{99.58}  &  \underline{99.71} \\
 
 \cline{2-8}
 & \multirow{5}{*}{GU$^2$-Net~\cite{zhu2021you}$^*$} & Imagenet~\cite{DBLP:conf/cvpr/DengDSLL009} & 0.683 & 96.13 & 97.98 &  \underline{98.77} & 99.39\\
 & & Zhu~\cite{zhu2021improving} & 0.684 \textcolor{red}{(+0.001)} & 96.17  & 97.95  & \textbf{98.79}  &  \underline{99.44} \\
 & & MOCHI~\cite{kalantidis2020hard} &  \underline{0.680} \textcolor{blue}{(-0.003)} &  \underline{96.20}  &  \underline{98.00}  & 98.75  & \textbf{99.44} \\
 & & Un-mix~\cite{shen2022mix} & 0.681 \textcolor{blue}{(-0.002)} & 96.17  & 97.95  & 98.74  & 99.42 \\
 & & ours & \textbf{0.671 \textcolor{blue}{(-0.012)}} & \textbf{96.21}  & \textbf{98.04}  & 98.72  & 99.43 \\
%  & & ours & 0.671 \textcolor{blue}{(-0.0?)} & 96.21  & 98.04  & 98.72  & 99.43 \\
 \cline{1-8}
 \multirow{5}{*}{One-shot}& \multirow{5}{*}{CC2D-S2~\cite{yao2021one}$^\#$} & Imagenet~\cite{DBLP:conf/cvpr/DengDSLL009} & 2.22 & 56.90 & 70.90 & 78.30 & 86.90 \\
 & & Zhu~\cite{zhu2021improving}  & 2.26 \textcolor{red}{(+0.04)} & 63.47  & 72.28  & 78.26  & 86.07 \\
 & & MOCHI~\cite{kalantidis2020hard} & 2.03 \textcolor{blue}{(-0.19)} & 67.76  & 76.02  & 81.78  & 88.88  \\
 & & Un-mix~\cite{shen2022mix} & \underline{1.98} \textcolor{blue}{(-0.24)} & \underline{68.35}  & \underline{77.02}  & \underline{82.73}  & \underline{89.57}  \\
& & ours  & \textbf{1.70 \textcolor{blue}{(-0.52)}} & \textbf{72.33}  & \textbf{80.04}  & \textbf{85.19}  & \textbf{92.49}  \\
%  & CC2D-Self & 1 (SCP~\cite{quan2021images}) & ? & 56.90 & 70.90 & 78.30 & 86.90 \\
\hline
\hline
\multicolumn{8}{c}{H\&N 3D Testset} \\ 
\hline
 \multirow{2}{*}{Type} & \multirow{2}{*}{Model} & \multirow{2}{*}{Pretrain} & \multicolumn{5}{c}{MRE ($\downarrow$) (mm)} \\
 \cline{4-8}
  &  & & Avg. & BS & MD & PL & PR \\ 
  \hline
  \multirow{5}{*}{One-shot} & \multirow{5}{*}{CC2D-S2~\cite{yao2021one}$^\#$} & Imagenet~\cite{DBLP:conf/cvpr/DengDSLL009} & \underline{10.78} & 7.70 &	16.30 &	\underline{8.65} & \underline{10.47} \\
  &  & Zhu~\cite{zhu2021improving}     & 14.50 \textcolor{red}{(+3.72)} & 7.47  &	18.17  &	16.96  & 15.43  \\
  &  & MOCHI~\cite{kalantidis2020hard} & 13.05 \textcolor{red}{(+2.27)} & \textbf{6.98}  &	\underline{15.27}  &	14.97  & 14.99  \\
  &  & Un-mix~\cite{shen2022mix}       & 11.75 \textcolor{red}{(+0.97)} & 7.31  &	\textbf{15.18}  &	12.32  & 12.22  \\
  & & ours & \textbf{10.10 \textcolor{blue}{(-0.68)}} & \underline{7.12}  &	15.32  &	\textbf{8.11}  &	\textbf{9.88}  \\
  \hline
\end{tabular}
\begin{tablenotes}
    \footnotesize
    \item[] *: re-running the official code with image size of $384\times384$. CC2D-S2~\cite{yao2021one}$^\#$: results of the 2nd stage of CC2D. Imagenet~\cite{DBLP:conf/cvpr/DengDSLL009}: loading weights pretrained with Imagenet for backbone. BS: Brainstem; MD: Mandible; PL: Left parotid; PR: Right parotid. 
\end{tablenotes}

\end{threeparttable}
\label{table:main}
\end{table*}

\subsection{Method summary}
Our methods can be concluded into 3 sequential steps as demonstrated in Figure~\ref{fig:pipeline} and Eqn.~(\ref{eq:L}): 
(1) Reducing $\rho_l$ by entropy-based sampling weight maps. 
(2) Estimating data augmentation parameters $A_l, A_m, A_h$ by mutual information; and
(3) Applying adaptive data augmentation.

\section{Experiment}
\subsection{Datasets and Settings}
\bheading{Cephalometric dataset:} It is a widely-used public dataset for cephalometric landmark detection, containing 400 radiographs, and is provided in IEEE ISBI 2015 Challenge~\cite{wang2016benchmark}. 
There are 19 landmarks of anatomical significance labeled by 2 expert doctors in each radiograph. We integrate the original two testsets into a combined testset.
% \footnote{Kaggle, Cephalometric X-Ray Landmarks Detection Challenge, \url{https://www.kaggle.com/jiahongqian/cephalometric-landmarks/discussion/133268}.}
% The average of the annotations by two doctors is set as the ground truth. The image size is $1935 \times 2400$ and the pixel spacing is 0.1mm. The dataset is split into 150 and 250 for training and testing respectively, referring to the official division. 
All analysis experiments are conducted on Cephalometric dataset. 

\bheading{Hand X-ray dataset:}
It is also a public dataset including 909 X-ray images of hands. The setting of this dataset follows~\cite{ref_scn}. The first 609 images are used for training and the rest for testing. 
% The image size varies among a small range, so all images are resized to 384$\times$384.

\bheading{H\&N 3D dataset:}
It is a 3D dataset constructed by Lei {\it et al.}~\cite{lei2021contrastive} with landmarks marked via segmentation masks from a mixed head and neck (H\&N) CT dataset.
% containing 98 volumes from two sources: 50 patients from StructSeg 2019~\footnote{\url{https://structseg2019.grand-challenge.org/Home/}}, 48 patients from MICCAI 2015 Head and Neck challenge~\cite{raudaschl2017evaluation}. We consider four organs that are annotated in all of them: brain stem, mandible, left and right parotid gland. The image is resampled to $3\times1\times1$ mm along axial, coronal, sagittal directions and cropped to remove the air background. The 98 volumes are split into 78 training and 20 testing samples. We pick the points with maximum or minimum values in x/y/z axis in a segmentation mask for each structure, yielding a total of 24 points as landmarks.

%\subsection{Settings}
\bheading{Metrics:} Following the official challenge~\cite{wang2016benchmark}, we use a mean radial error (MRE) to measure the Euclidean distance between prediction and ground truth, and successful detection rate (SDR) in four radii (2mm, 2.5mm, 3mm, and 4mm). 

\bheading{Model:} We build our encoder following the SSL module of CC2D~\cite{yao2021one}. All model and training settings are the same as \cite{yao2021one}. In addition, the one-shot template used for evaluation is selected by \cite{quan2021images} in an unsupervised style. 

\bheading{Implementation details:}
All training and testing images are resized to $384 \times 384$. We intuitively define $\{0, 2, 4\}$ as thresholds of low-, medium-, high-info groups.
$\gamma = 0.3$ on Cephalometric dataset and $\gamma = 0.2$ in Hand X-ray and H\&N 3D datasets. We set $\hat{\alpha}$ to 1 for all three datasets. The patch size $k$ for estimating IIE is set to 10 for Cephalometric and Hand X-ray datasets, and 5 for H\&N 3D dataset. The CT values in H\&N 3D dataset are clipped into $[-200, 400]$.

\begin{figure*}[!h]
\begin{minipage}{0.36\linewidth}
 \centering
 \includegraphics[width=\linewidth]{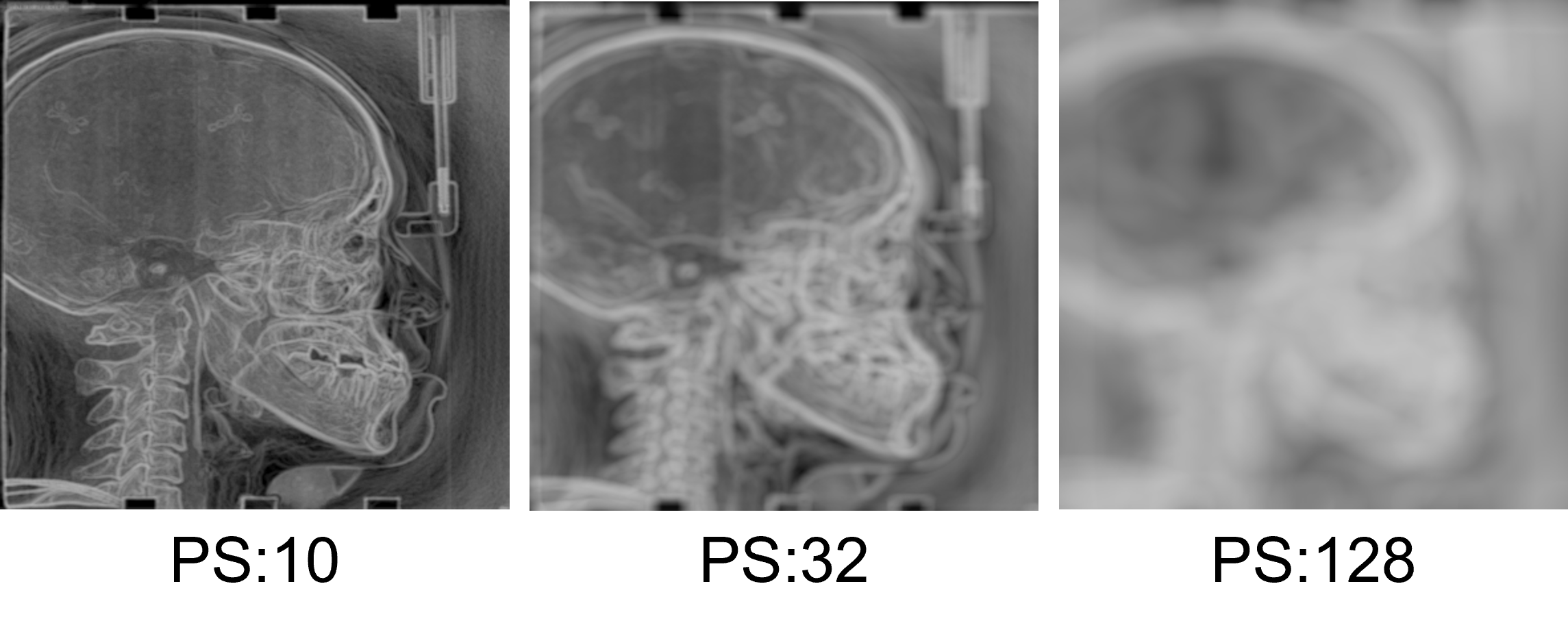}
\end{minipage}
\begin{minipage}{0.43\linewidth}
 \centering
 \includegraphics[width=0.9\linewidth]{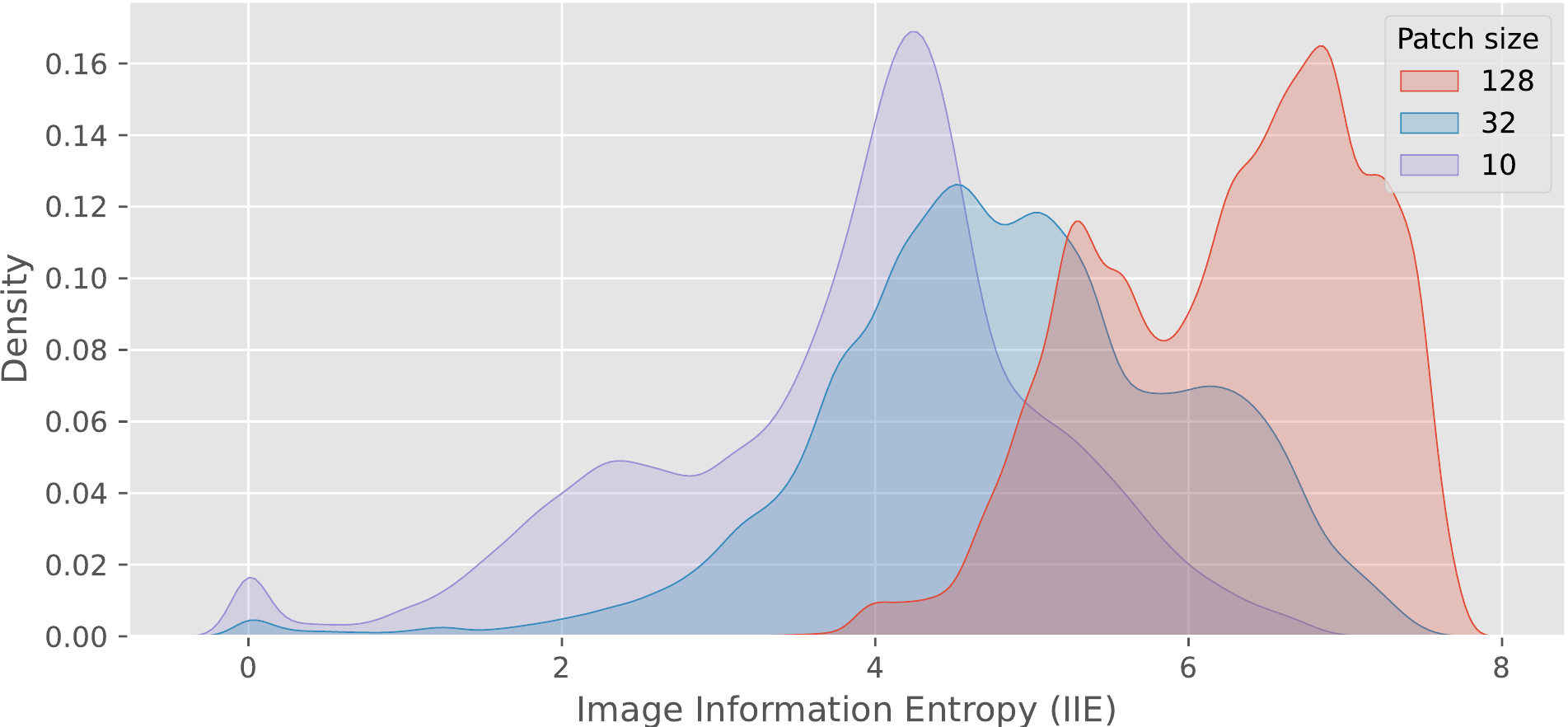}
\end{minipage}
\begin{minipage}{0.2\linewidth}
    \centering
    \begin{tabular}{l|c}
    \multicolumn{2}{c}{Analysis of PS} \\
    \hline
        PS  & MRE ($\downarrow$)  \\
        \hline
        10 & \textbf{2.34}   \\
        \hline
        32 & 2.37 \\
        \hline 
        128 & 2.39 \\
    \hline
    \end{tabular}
\end{minipage}
\caption{Left: the IIE maps generated by different sizes of image patches; Middle: the probability density of IIE with different patch sizes; and Right: results with IIE maps of different patch sizes (PS).}
\label{fig:entr_ps}
\end{figure*}

\subsection{Main results}
To directly measure the performance of our encoder, we quantitatively compare our method and other latest methods, Zhu~\cite{zhu2021improving}, MOCHI~\cite{kalantidis2020hard} and Un-mix~\cite{shen2022mix} on the local feature matching problem (Section~\ref{sec:task}). Results are listed on Table~\ref{table:main2}. It should be noted that, we use the augmentation parameters estimated for high-info pixels by our method in Zhu~\cite{zhu2021improving}, MOCHI~\cite{kalantidis2020hard} and Un-mix~\cite{shen2022mix} for fair comparison. Our estimated augmentation parameters perform much better than the original parameters.
Because our CL model consists of two independent encoders that do not share weights and are optimized by their own gradients, cross-fusing the features from two different projection spaces in \cite{zhu2021improving} leads to performance degradation. MOCHI~\cite{kalantidis2020hard} leverages mixed negative features and hard example mining to concentrate more on negative views, but they ignore the influence of positive views. Un-mix~\cite{shen2022mix} creates views by linear interpolation on the image level, lacking pixel-level augmentation.
Compared with these methods, we increase the diversity at the pixel level with adaptive effective augmentation.
Our method {\bf consistently} outperforms other methods and greatly improves the baseline model by 0.57mm in MRE (from 2.91mm to 2.34mm, {\bf a 19.6\% reduction}).

We also load these SSL weights as pretrained weights on several supervised models~\cite{ref_scn,zhu2021you,McCouat_2022_CVPR} as well as a one-shot models~\cite{yao2021one} and evaluate them on Cephalometric~\cite{wang2016benchmark} and Hand-Xray datasets. It should be noted that CC2D consists of two stages, self- and semi-supervised learning modules. We apply our pre-trained model as the stage one model and then finetune on the semi-supervised model. As shown in Table~\ref{table:main}, all models benefit from most pretrained models on MRE. In particular, models based on our SSL pretrained model achieve the largest improvement. Such an improvement is more prominent for one-shot model CC2D-S2. For example, when compared with the one-shot CC2D-S2 model that is based on a pretrained model from ImageNet, we reduce the MRE by 0.54mm (from 2.85mm to 2.21mm, \textbf{a 22.4\% reduction}) in the Cephalometric testset and by 0.52mm (from 2.22mm to 1.70mm, \textbf{a 23.4\% reduction}) in the Hand X-ray testset. In addition, some SDR values degrade because the pretrained model tends to optimize the bad cases and pay less attention to the good ones. 
% Figure~\ref{fig:hard_case} demonstrates such issue that the original model make bad predictions for hard cases, but model with our pretrained model are more robust to the hard cases.

\subsection{Ablation study}
\bheading{Components:}
Table~\ref{table:abla} demonstrates the impact of each module. 
As shown in Table~\ref{table:abla}, both IIE mapping and adaptive augmentation with parameter estimation bring huge improvements to SSL training due to effective noise reduction and successfully unleashing the potential of each pixel. The IIE map greatly improves the baseline by 0.45mm in MRE, and the adaptive augmentation also helps to further improve the model by 0.12mm in MRE. 

\bheading{Hyperparameter $\hat{\alpha}$:}
Figure~\ref{fig:alpha} demonstrates that $\hat{\alpha}$ close to 1 is be a good choice. When $\alpha$ increases from 1, the model degenerates slowly; when $\alpha$ decreases from 0.8, the model degenerates greatly. A too low $\hat{\alpha}$ is not recommended.

\bheading{Generation of IIE maps:}
The generation of IIE maps is relevant to our method. Compared with a patch size of 10 ({ default} in our method), larger patch sizes (32 and 128) lead to a smoother entropy graph (Figure~\ref{fig:entr_ps}(left)) and including more information increases the pixel's entropy and results in a narrow range of IIE, thereby failing to capture local details important for landmark detection (Figure~\ref{fig:entr_ps}(middle)). The table in Figure~\ref{fig:entr_ps}(right) also demonstrates that IIE maps with larger patch sizes perform worse. The model degrades in MRE when the patch size increases from 10 to 32.

\section{Conclusion} 
We propose an information-guided pixel augmentation strategy to boost pixel-wise CL. We first classify pixels into three categories, namely low-, medium-, and high-informative, based on the information quantity the pixel contains. 
Inspired by the ``InfoMin" principle, we then design separate strategies for each category 
including: (1) sampling pixels with entropy-based weight maps; (2) estimating augmentation parameters by mutual information; (3) applying adaptive augmentation. 
Numerous experiments validate that our information-guided pixel augmentation strategy succeeds in encoding more discriminative representations and outperforms other competitive approaches in unsupervised local feature matching. Additionally, supervised models can also be enhanced by our pretrained model. we plan to investigate better estimation strategy to further close the gap between optimal augmentation parameters and our estimates. %, and adaptive augmentation can be built in a better form theoretically. 

%%%%%%%%% REFERENCES

\newpage

{\small
\bibliographystyle{ieee_fullname}
\bibliography{egbib}
}

\end{document}